\makeatletter 
\renewcommand\@biblabel[1]{#1.} 
\makeatother

\documentclass[runningheads]{llncs}
\usepackage[misc]{ifsym}
\usepackage{graphicx}
\usepackage{subfigure} 
\usepackage{amssymb}
\usepackage{multirow}
\usepackage{amsmath, bm}
\usepackage{hyperref}
\hypersetup{
    colorlinks=true,
    linkcolor=black,
    filecolor=black,      
    urlcolor=black,
    citecolor=black,
}
\begin{document}
\title{IFR: Iterative Fusion Based Recognizer For Low Quality Scene Text Recognition}
\titlerunning{IFR}
\author{Zhiwei Jia \and Shugong Xu\textsuperscript{(\Letter)} \and Shiyi Mu \and Yue Tao \and Shan Cao \and Zhiyong Chen }
\authorrunning{Jia et al.}

\institute{Shanghai Institute for Advanced Communication and Data Science,\\
Shanghai University, Shanghai, 200444, China\\
\email{\{zhiwei.jia, shugong, mushiyi, yue\_tao, cshan, bicbrv\_g\}@shu.edu.cn}}

\maketitle       
\begin{abstract}
Although recent works based on deep learning have made progress in improving recognition accuracy on scene text recognition, how to handle low-quality text images in end-to-end deep networks remains a research challenge. In this paper, we propose an \textbf{I}terative \textbf{F}usion based \textbf{R}ecognizer (\textbf{IFR}) for low quality scene text recognition, taking advantage of refined text images input  and robust feature representation. IFR contains two branches which focus on scene text recognition and low quality scene text image recovery respectively. We utilize an iterative collaboration between two branches, which can effectively alleviate the impact of low quality input. A feature fusion module is proposed to strengthen the feature representation of the two branches, where the features from the \textbf{R}ecognizer are \textbf{F}used with image \textbf{R}estoration branch, referred to as \textbf{RRF}. Without changing the recognition network structure, extensive quantitative and qualitative experimental results show that the proposed method significantly outperforms the baseline methods in boosting the recognition accuracy of benchmark datasets and low resolution images in TextZoom dataset.
\keywords{Scene Text Recognition \and Iterative Collaboration \and Feature Fusion.}
\end{abstract}
\section{Introduction}
In recent years, scene text recognition (STR) has attracted much attention of the computer vision community. STR aims to recognize the text in the scene images, which is an important part of the downstream task, such as license plate recognition\cite{wu2018many}, receipts key information extraction\cite{wang2021towards}, etc. Recent works based on deep learning have succeeded in improving recognition accuracy on clear text images. Benefiting from the development of sequence-to-sequence learning, STR methods can be roughly divided into two major techniques\cite{long2021scene}, Connectionist Temporal Classification \cite{shi2016end,wan20192d} and Attention mechanism\cite{li2019show,luo2019moran,shi2018aster,wang2020decoupled}.

\begin{figure}[htbp]
\centering
\subfigure[]{
\includegraphics[width=0.45\textwidth]{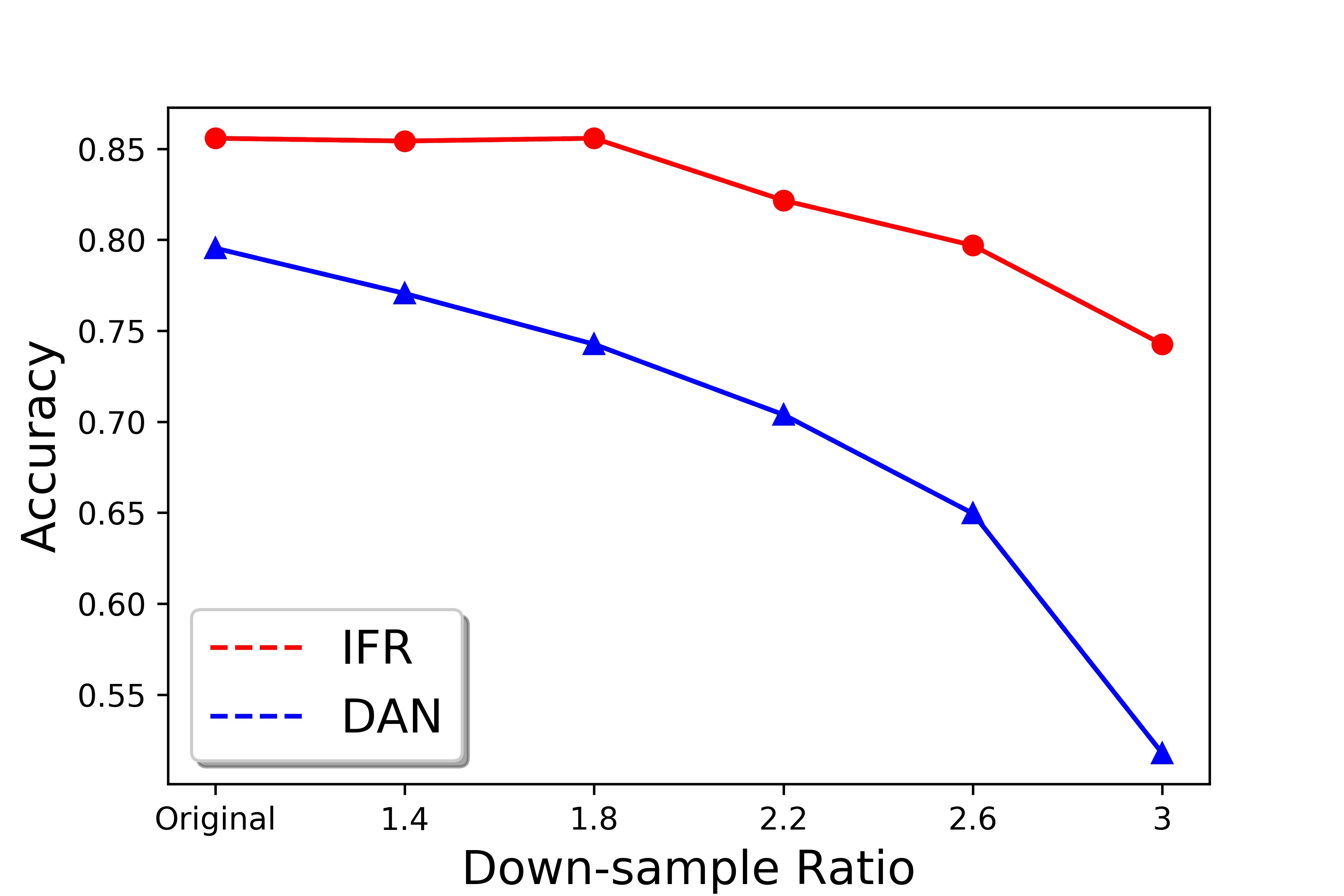} 
}
\quad
\subfigure[]{
\includegraphics[width=0.45\textwidth]{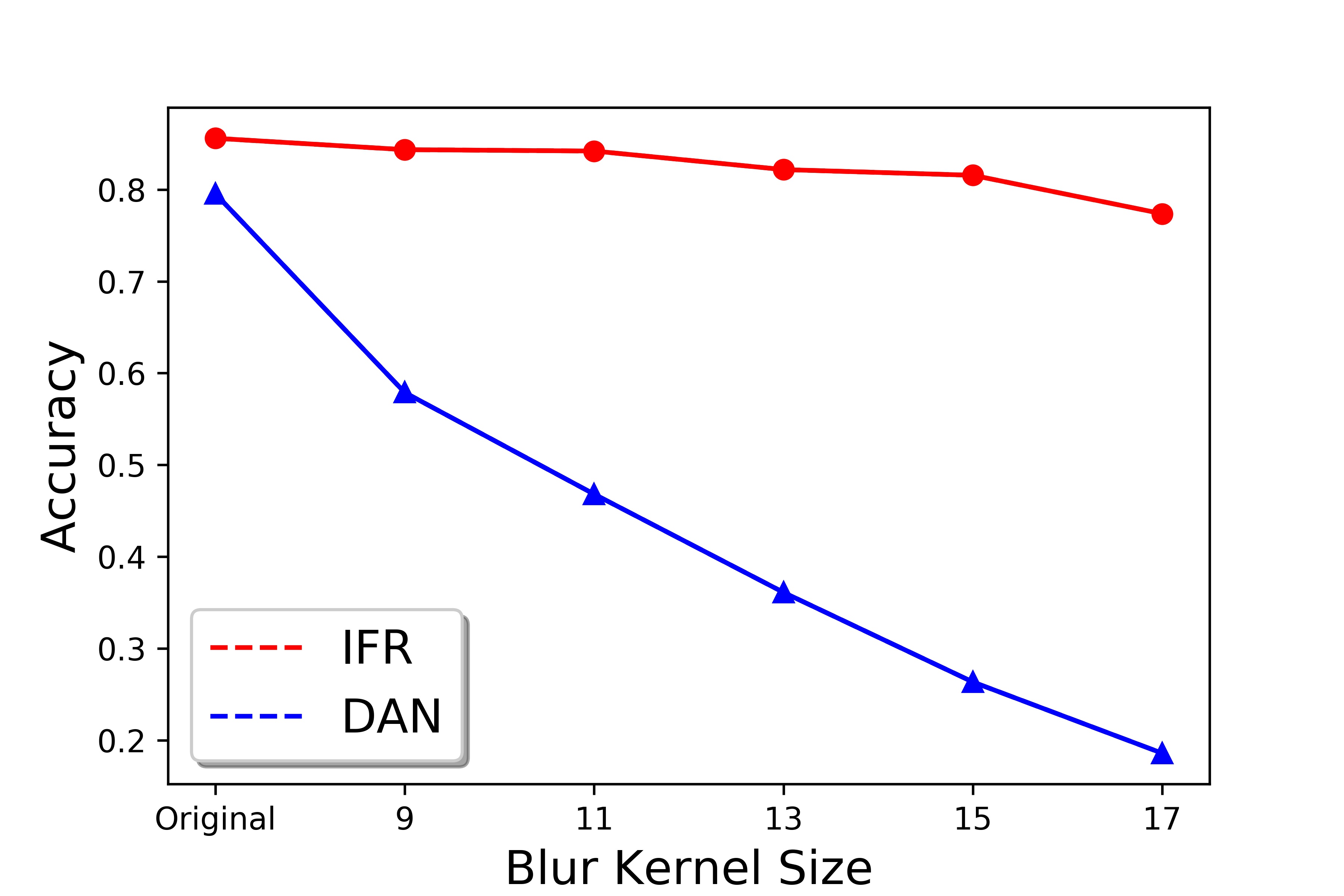}
}
\quad
\subfigure[]{
\includegraphics[width=\textwidth]{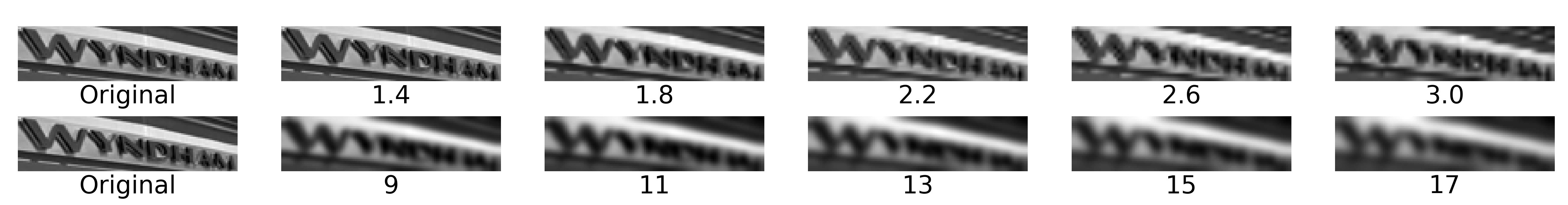}
}
\caption{Accuracy comparison on SVTP\cite{quy2013recognizing} dataset. (a-b)Compared to state-of-the-art DAN\cite{wang2020decoupled} (in blue), our IFR model (in red) is shown to achieve robust results for different image degradation methods. (c)Visualization results of different degrees of down-sample ratio(up) and blur kernel size(down).}
\label{degradation}
\end{figure}

\begin{figure}[htbp]
\centering
\subfigure[]{
\includegraphics[width=0.3\textwidth]{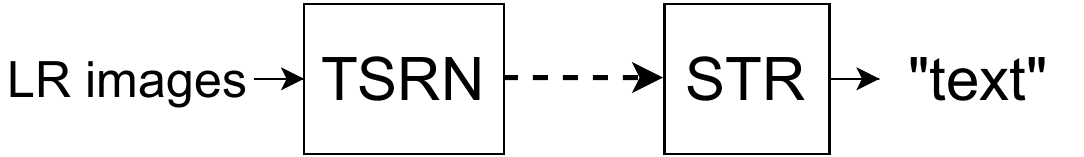}
}
\subfigure[]{
\includegraphics[width=0.3\textwidth]{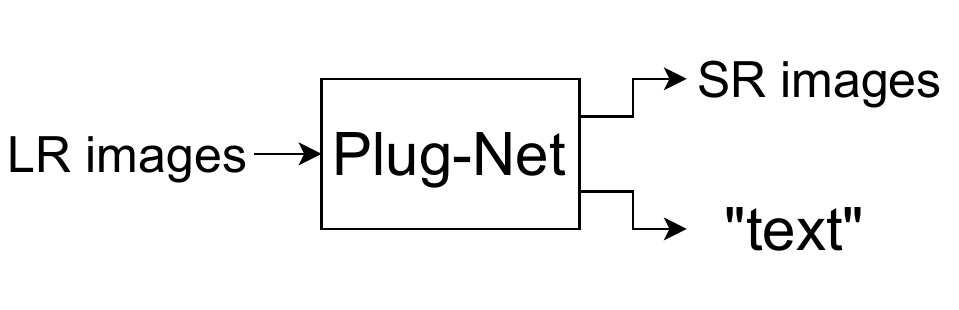}
}
\subfigure[]{
\includegraphics[width=0.3\textwidth]{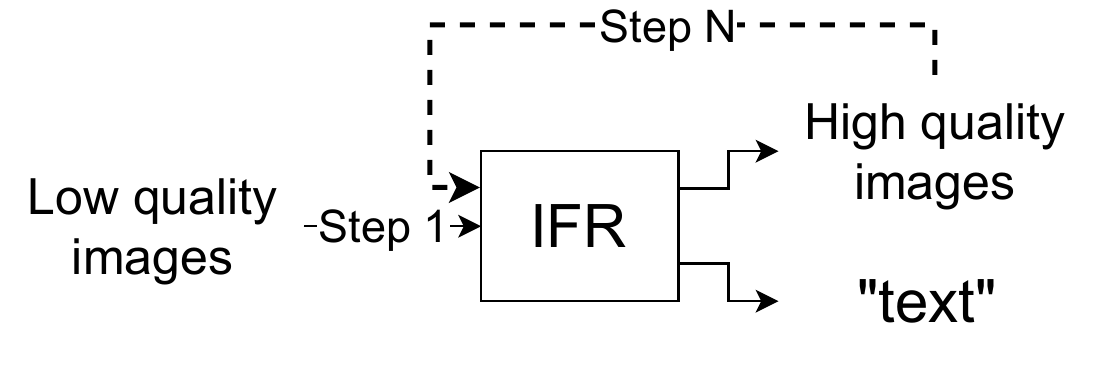}
}
\caption{(a) Pipeline of TSRN. The two network are trained separately and the dotted line represent inference. (b) Pipeline of Plug--Net which containing two branches. (c) Our iterative fusion recognizer IFR.}
\label{net_pipeline}
\end{figure}

Previous works focus on texts in natural\cite{shi2016end,cheng2017focusing} and curve scenes\cite{shi2018aster,li2019show,luo2019moran}, which prove the outstanding performance on clear images. However, as shown in Fig. \ref{degradation}, both of the methods are facing a problem that their performances drop sharply when text image quality is poor, including low--resolution, blurred, low contrast, noisy, etc. Therefore, how to generalize the recognizer to both high and low quality text images is still a research challenge. The biggest problem is that it is hard to extract robust semantic information due to the lack of sufficient text region visual details. Recent works have noticed this problem. TSRN\cite{wang2020scene}(Fig. \ref{net_pipeline}a) introduced super--resolution (SR) methods as a pre--processing procedure before recognizer and show good performance on the scene text SR dataset TextZoom, which is the first real paired scene text image super-resolution dataset. The limitation is that TSRN can only handle the problem of low--resolution text recognition. A more reasonable way is image restoration(IR)\cite{zamir2021multi}, which is the task of recovering a clean image from its degraded version. However, compared to single image restoration, text image restoration only considers text-level features instead of complex scenes. So it is necessary to share features between STR and IR. In this way, the feature of the recognizer backbone can be strong prior knowledge for restoration. Although Plut--Net\cite{eccv2020plugnet}(Fig. \ref{net_pipeline}b) proposed an end-to-end trainable scene text recognizer together with a pluggable super--resolution unit, the super--resolved images from the IR branch are not utilized to enhance the recognition accuracy. As the IR branch restores higher quality images, they can be used as new inputs into the recognition branch to get more accurate results. 

Inspired by these works, we propose an iterative fusion based recognizer as shown in Fig. \ref{net_pipeline}c. IFR contains two networks focusing on scene text recognition and low--quality scene text image restoration, respectively. Benefit from the IR branch, IFR uses refined text images as new inputs for better recognition results in the iterative collaboration manner, which utilizes image restoration knowledge in scene text recognition. In each step, previous outputs of IR branch are fed into the STR branch in the following step. Different from previous methods, the text image restoration and recognition processes facilitate each other progressively. A fusion module RRF is designed to combine the image shallow feature and the STR backbone feature. In this way, the IR branch can get more semantic information to refine the detail of text region instead of background and noise. In return, feature sharing enhances the feature representation of different quality text images. We also claim that our proposed RRF and collaboration manner can boost the performance on standard benchmarks and low--resolution dataset TextZoom without modifying the recognizer. Additionally, only exploiting synthetic augmented training data can achieve the comparable results on TextZoom.  

The contributions of this work are as follow:
\begin{itemize}
\item[*] Firstly, an iterative fusion based recognizer IFR are proposed for scene text recognition which contains two branches focusing on scene text recognition and low--quality scene text image recovery, respectively. An iterative collaboration are then proposed between two branches, which can effectively alleviate the impact of low--quality input. 

\item[*] Secondly, a fusion module RRF is designed to strengthen the feature representation of the two tasks. Benefited from this module, the recognition branch can enhance the feature extraction for recognizing low-quality text images.

\item[*] Thirdly, without modifying the recognizer, quantitative and qualitative experimental results show the proposed method outperforms the baseline methods in boosting the recognition accuracy of benchmark datasets and LR images in TextZoom dataset.
\end{itemize}

\section{Related works}
\subsubsection{Text Recognition.} Early work used a bottom-up approach which relied on low--level features, such as histogram of oriented gradients descriptors\cite{wang2011end}, connected components\cite{neumann2012real}, etc. Recently, deep learning based methods have achieved remarkable progress in various computer vision tasks including scene text recognition. STR methods can roughly divided into two major techniques\cite{long2021scene}: Connectionist Temporal Classification\cite{graves2006connectionist} and Attention mechanism. CRNN\cite{shi2016end} integrates feature extraction, sequence modeling, and transcription into a unified framework. CTC is used to translate the per-frame predictions into the final label sequence. As the attention mechanism was wildly used in improving the performance of natural language process systems, an increasing number of recognition approaches based on the attention mechanism have achieved significant improvements. ASTER\cite{shi2018aster} proposed a text recognition system which combined a Spatial Transformer Network (STN)\cite{jaderberg2015spatial} and an one dimension attention-based sequence recognition network. DAN\cite{wang2020decoupled} propose a decoupled attention network, which decouples the alignment operation from using historical decoding results to solve the attention drift problem. In this work, we choose DAN as our baseline recognizer to boost the recognition accuracy of the low-quality images.
\subsubsection{Low-Quality Scene Text Image Recognition.}
Recent works have noticed the text image degradation problem. TSRN introduced the first real paired scene text SR dataset TextZoom and proposed a new text SR method as a pre--processing procedure before recognition methods. Plug--net proposed an end-to-end trainable scene text recognizer together with a pluggable super-resolution unit for a more robust feature representation.

These works have achieved notable success by super resolution module, but they treat the two networks as an independent task. While recursive networks and feature fusion promote the development of image restoration\cite{zamir2021multi}, a few methods have employed their generative power in low-quality text recognition. To our best knowledge, this is the first attempt to integrate text recognition and image restoration iteratively into a single end-to-end trainable network. 

\begin{figure}[ht]
\centering
\includegraphics[width=\textwidth]{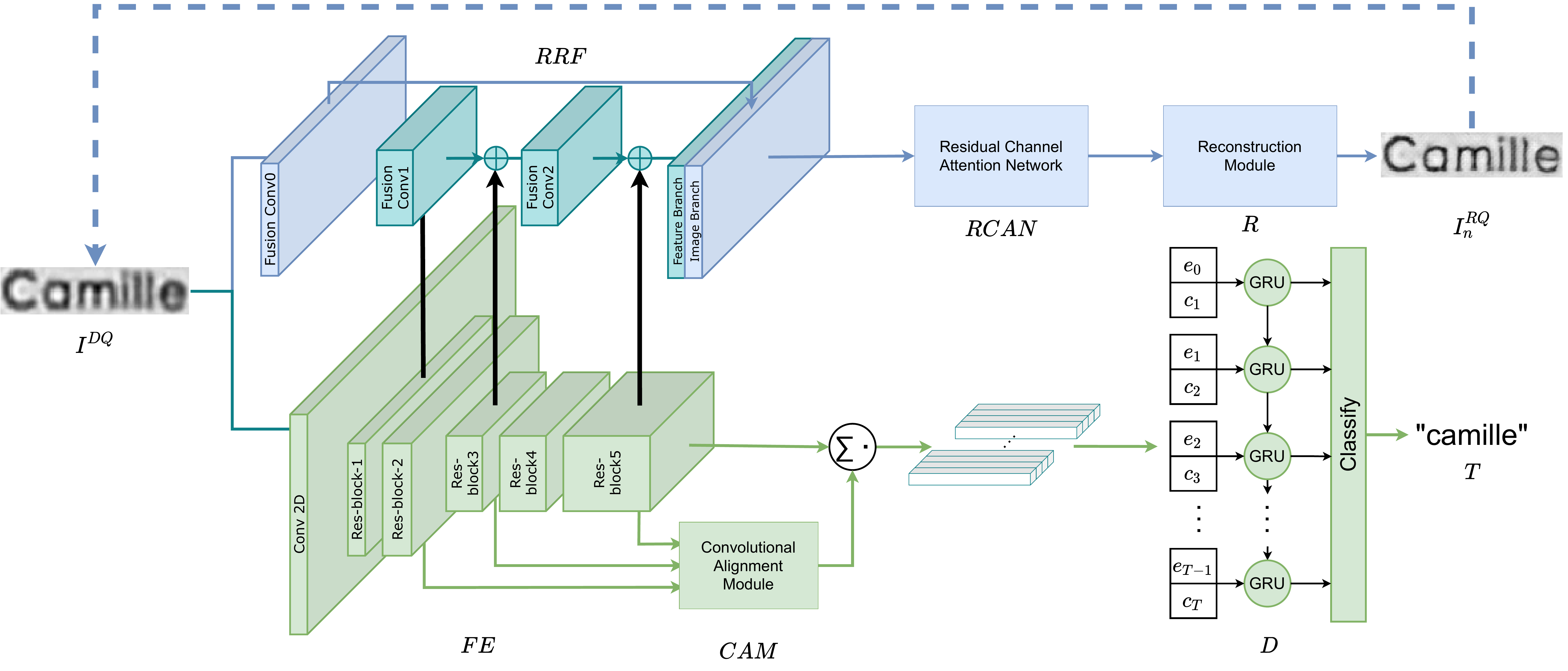}
\caption{Overall framework of IFR.}
\label{network}
\end{figure}

\section{Method}
In this section, we present our proposed method IFR in detail. As shown in Fig. \ref{network}, we aim to recognize the text of the input degraded quality text images $\mathbf{I}^{DQ}$ and get the high-quality result $\mathbf{I}^{HQ}$. We design a deep iterative collaboration network that estimates high-quality text images and recognizes texts iteratively and progressively between the recognizer and IR process. In order to enhance the collaboration, we design a novel fusion module RRF that integrates two sources of information effectively. 

\subsection{Iterative Collaboration}
The previous methods are very sensitive to image quality as low-quality images may lose enough information for recognition. Therefore, our method alleviates this issue by an iterative fusion IFR as shown in Fig. \ref{network}. In this framework, text image recovery and recognition are performed simultaneously and recursively. We can obtain more accurate recognition result if the text image is restored well. Both processes can enhance each other and achieve better performance progressively. Finally, we can get accurate recognition results and HQ text images with enough steps.

The recognition branch uses DAN as the baseline, which includes a CNN-based feature encoder FE, a convolutional alignment module CAM, and a decoupled text decoder D. The CAM takes multi-scale visual features from the feature encoder as input and generates attention maps with a fully convolutional network\cite{long2015fully} in channel-wise manner. Compare with other attention decoders, DAN avoids the accumulation and propagation of decoding errors. The IR branch contains a fusion module RRF, a residual channel attention network RCAN\cite{zhang2018image} and a reconstruction module R. RCAN contains two residual groups, each of which has two residual channel attention blocks. Residual groups and long skip connection allow the main parts of the network to focus on more informative components of the low-quality features. Channel attention extracts the channel statistic among channels to further enhance the discriminative ability of the network.

For the first step, the recognition branch extracts the multi-scale image features and decodes the text result by using the degraded quality text images, denoted as $\{\mathbf{f}_{1}^T\}$ and $\mathbf{T}_1$, respectively. Besides, the image features are fed into the fusion module with the input image. Therefore, the first step can be formulated by:
\begin{equation}
\{\mathbf{f}_{1}^T\} = FE(\mathbf{I}^{DQ}), 
\end{equation}
\begin{equation}
\mathbf{T}_1 = D(CAM(\{\mathbf{f}_{1}^T\}), \{\mathbf{f}_{1}^T\}),
\end{equation}
\begin{equation}
\mathbf{f}_1^I = RRF(\mathbf{I}^{DQ}, \{\mathbf{f}_{1}^T\}),
\end{equation}
\begin{equation}
\mathbf{I}_1^{RQ} = R(RCAN(\mathbf{f}_1^I)),
\end{equation}
For the $n_{th}$ step where $n=2, ..., N$, the difference is the input in feature encoder FE and fusion module RRF is the reconstructed picture $\mathbf{I}_{n-1}^{RQ}$ from the previous step $n-1$, as follows:
\begin{equation}
\{\mathbf{f}_{n}^T\} = FE(\mathbf{I}_{n-1}^{RQ}), 
\end{equation}
\begin{equation}
\mathbf{T}_n = D(CAM(\{\mathbf{f}_{n}^T\}), \{\mathbf{f}_{n}^T\}),
\end{equation}
\begin{equation}
\mathbf{f}_n^I = RRF(\mathbf{I}_{n-1}^{RQ}, \{\mathbf{f}_{n}^T\}),
\end{equation}
\begin{equation}
\mathbf{I}_n^{RQ} = R(RCAN(\mathbf{f}_n^I)),
\end{equation}
After N iterations, we get the recognition results $\{\mathbf{T}_n\}_{n=1}^N$ and reconstruction result $\{\mathbf{I}_n^{RQ}\}_{n=1}^N$. In order to optimize the output of each iteration, we calculate the loss for each step of the output. In this way, the output results are gradually optimized through supervision. 
\subsection{Fusion Module RRF}
Previous works considered text image restoration as independent tasks to improve the image quality or multi-task to obtain a more robust feature representation. Our method exploits the merits of both designs. In our method, a fusion module is designed to strengthen the feature representation of the two tasks, where the features from the recognizer are fused with IR branch. As shown in Fig. \ref{network}, the fusion module contains a 3 × 3 convolutional layer for the image shallow feature extraction and a cascade convolutional layer to generate the multi-scale fusion features. The inputs are degraded quality text images $\mathbf{I}^{DQ}$ at the first step or $\mathbf{I}_n^{RQ}$ from the last step, and visual features $\{\mathbf{f}_{n}^T\}$ from the feature encoder. These multi-scale features are first encoded by cascade convolutional layers then concatenate with image shallow feature as output. The proposed RRF has several merits. First, the STR branch can extract more robust features for recognizing low-quality text images. Second, the multi-scale features of the FE help to enrich the semantic features of the IR branch.

\subsection{Loss Functions}
Here, IFR is trained end-to-end using the cross-entropy loss $L_{rec}$ and the pixel-wise loss $L_{pixel}$. the $L_{rec}$ is defined as follow:
\begin{equation}
L_{rec} = - \frac{1}{2} \sum_{i=1}^T (\log p_{LR}(y_t|\mathbf{I})+\log p_{RL}(y_t|\mathbf{I}))
\end{equation}
where $y_t$ is the ground--truth text represented by a character sequence. $p_{LR}$ and $p_{RL}$ are bidirectional decoder distributions, respectively. The pixel-wise loss function is defined as follow:
\begin{equation}
L_{pixel} = \frac{1}{N*W*H} \sum_{n=1}^N \sum_{i=1}^W \sum_{j=1}^H ||\mathbf{I}^{HQ} - \mathbf{I}_n^{RQ}||
\end{equation}
W and H refer to the width and height of the input image. The model is optimized by minimizing the following overall objective function:
\begin{equation}
L= L_{rec} + \lambda L_{pixel}
\end{equation}
Note that the gradients can be back-propagated to both the STR and IR in a recursive manner. The STR can be supervised by not only the recognition loss but also by the revision of image restoration loss through the fusion module.

\subsection{Paired Training Data generate}
For fair comparison, our model is trained on the MJSynth MJ~\cite{jaderberg2014synthetic,jaderberg2016reading} and SynthText ST~\cite{gupta2016synthetic}. As we all know, the smaller the domain gap between the training dataset and the real scene, the better the performance of the test dataset. TSRN experiment shows that fine-tune ASTER on TextZoom training set can improve the accuracy of the TextZoom test set but harm the performance of other high-quality benchmarks. In our work, data augmentation is a key method to generate paired high and low-quality training text images. Owing to the good quality of the two synthetic datasets, we use random data augmentation, Gaussian kernel and down-up sampling, to generate paired training data. Inspired by Plug--Net, these methods randomly generate paired data. The degraded text images $\mathbf{I}^{DQ}$ are the input at the first step and $\mathbf{I}^{HQ}$ provide ground--truth supervisory signals useful for the progressive image restoration at each step. Different from other augment methods, The degraded images contain both clear and different degrees low-quality text images as the random strategy, which enables IR branch to learn not only “how” but also “when” to restore a text image. 

\section{Experiments}
\subsection{Datasets and Implementation Details}
Eight standard benchmarks include ICDAR 2003 (IC03)~\cite{lucas2005icdar}, ICDAR 2013 (IC13)~\cite{karatzas2013icdar}, ICDAR 2015 (IC15)~\cite{karatzas2015icdar}, IIIT5K (IIIT)~\cite{mishra2012scene}, Street View Text (SVT)~\cite{wang2011end}, Street View Text-Perspective (SVTP)~\cite{quy2013recognizing}, CUTE80 (CUTE)~\cite{risnumawan2014robust} and TextZoom are as the testing datasets. Details of these datasets can be found in the previous works~\cite{wang2020decoupled}. In addition, TextZoom dataset is divided into three testing subsets\cite{wang2020scene} by difficulty and a training set. 

We adopt an opensource implementation of \href{https://github.com/Wang-Tianwei/Decoupled-attention-network}{DAN} and reproduced bidirectional decoding according to\cite{wang2020decoupled}. The STR and IR model dimension C are set to 512 and 64 throughout respectively. Balanced factor $\lambda$ is set to 10. Ground--truths are directly resized to 32 × 128 greyscale images. Then the augment functions are used to generate the paired degraded data randomly. The range of blur kernel size is 9 to 17. The down-up sample ratio is in the range of 1 to 3. The model is trained by Adadelta optimizer. The initial learning rate is 1 and is decayed to 0.1 and 0.01 respectively after 4 and 5 epochs. Recognition results are evaluated with accuracy. IR results are evaluated with PSNR and SSIM. They are computed on the greyscale space. Our experiments are implemented on Pytorch with NVIDIA RTX 2080Ti GPUs.

\subsection{Ablation Study}

\subsubsection{Effectiveness of Iterative Learning}
In our work, we use the execution manner of iterative collaboration between STR and IR branches. To better show the influence of the proposed iterative learning scheme, we evaluate the recognition accuracy and quality of the IR outputs on different steps. The performance on TextZoom subsets is given in Table \ref{tab3}, where the testing steps are set to 1 to 4. As we can see from the results, iterating the IFR three times can achieve a significant improvement especially on the second step. Specifically, there are little gains on the third and fourth steps. As mentioned above, we use PSNR and SSIM as image recovery measurement metrics. From the first to fourth step, SSIM and PSNR get better progressively on most of the dataset. Therefore, the comparison proves that our method is able to achieve progressively better image quality and recognition accuracy simultaneously.

\begin{table*}[h]
\caption{Ablation study of iterative steps on TextZoom subsets. Step means the iteration times.}
\vspace{-0.6cm}
\label{tab3}
\begin{center}
\setlength
\tabcolsep{2pt}
\begin{tabular}{|c|ccc|ccc|ccc|}
\hline
\multirow{2}{*}{Step} & \multicolumn{3}{c|}{Hard}                         & \multicolumn{3}{c|}{Medium}                        & \multicolumn{3}{c|}{Easy}                         \\
                      & Accuracy       & SSIM            & PSNR           & Accuracy       & SSIM            & PSNR           & Accuracy       & SSIM            & PSNR           \\\hline
1                     & 45.87          & 0.6867          & 19.75          & 62.03          & 0.6504          & \textbf{19.50} & 78.43          & 0.8127          & 23.23          \\
2                     & 52.51          & 0.6940          & 19.82          & 69.31          & 0.6536          & 19.49          & 83.15          & 0.8280          & 23.57          \\
3                     & 52.77          & 0.7005          & \textbf{19.87} & 69.56          & 0.6562          & 19.45          & 83.36          & 0.8389          & 23.84          \\
4                     & \textbf{52.85}          & \textbf{0.7037} & 19.86          & \textbf{69.64}          & \textbf{0.6568} & 19.38          & \textbf{83.43} & \textbf{0.8445} & \textbf{23.97} \\ \hline
\end{tabular}
\end{center}
\vspace{-0.6cm}
\end{table*}

\begin{figure}[ht]
\vspace{-0.6cm}
\centering
\includegraphics[width=0.5\textwidth]{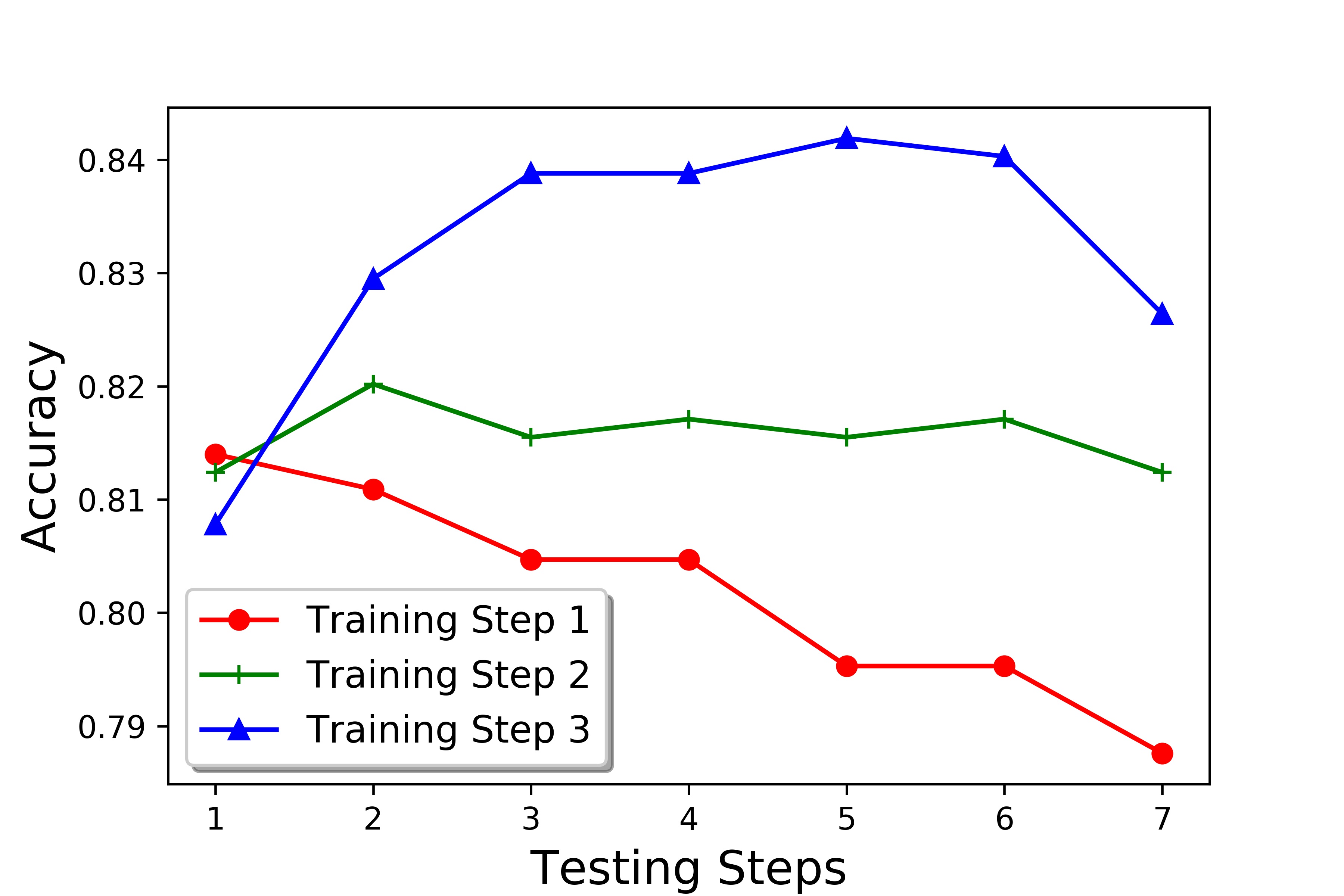}
\caption{Accuracy of iterating steps in training and testing phase on SVTP.}
\vspace{-0.4cm}
\label{iteration_steps}
\vspace{-0.4cm}
\end{figure}

We further explore the difference of iteration steps between training and testing. Increase the number of iteration steps at the training stage will increase the memory usage and training time. The same as the testing stage. The average accuracy on SVT--P in Fig. \ref{iteration_steps} suggests that: 1) training without applying iterative collaboration, the reconstructed images will harm the accuracy in STR branch when testing; 2) iterating during the training phase is helpful, as it provides recovery training images for STR branch; 3) the accuracy improves significantly in the second step and slowly improves in the next few steps. Therefore testing with a big iteration step is unnecessary. 

\begin{table*}
\caption{Ablation study of training data augmention and fusion module RRF. DAN$_{RP}$ represents the result we reproduced. Subscript numbers indicate the testing steps.}
\vspace{-0.4cm}
\label{tab1}
\begin{center}
\begin{tabular} {|c|c|c|c|c c c c c c c|c c c|}
\hline
\multirow{2}{*}{Method} & \multirow{2}{*}{Step} & \multirow{2}{*}{Aug} & \multirow{2}{*}{RRF} &  IIIT & SVT & IC03 & IC13 & IC15 & SVTP & CUTE& Hard & Medium & Easy\\
& & & & 3000 & 647 & 867 & 1015 & 2077 & 645 & 288 & 1175 & 1209 & 1442\\
\hline
DAN$_{RP}$ & 1 & \texttimes & \texttimes & 93.9 & 88.7 & 94.5 & 92.6 & 74.2 & 79.5 & \textbf{82.3} & 33.2 & 45.2 & 62.3\\
DAN$_{RP}$ & 1 & \checkmark & \texttimes & 93.9 & \textbf{89.2} & 94.7 & \textbf{93.7} & \textbf{76.5} & \textbf{82.1} & 80.6 & 36.3 & 50.0 & 70.3\\
IFR$_1$ & 1 & \checkmark & \checkmark & \textbf{94.5} & \textbf{89.2} &\textbf{94.8} & 93.4 & 76.0 & 81.3 & 80.6 & \textbf{37.9} & \textbf{50.8} & \textbf{71.2}\\
\hline
IFR$_2$ & 2 & \checkmark & \texttimes & 94.0 & 89.8 & 94.7 & 93.6 & 76.0 & 81.7 & \textbf{82.3} & \textbf{40.0} & 50.8 & 74.5\\
IFR$_2$ & 2 & \checkmark & \checkmark & \textbf{94.2} & \textbf{90.3} & \textbf{95.5} & \textbf{93.9} & \textbf{77.7} & \textbf{82.3} & 81.9 & 39.6 & \textbf{53.3} & \textbf{76.4}\\
\hline
\end{tabular}
\end{center}
\vspace{-0.4cm}
\end{table*}

\subsubsection{Effectiveness of Fusion Module}
Firstly, we discuss the performance from the training data. Different from the baseline method DAN, our work uses degraded images as input. For a fair comparison, we train the recognition branch DAN with original and augmentation datasets. Tabel \ref{tab1} shows the result of two recognition models in seven text recognition benchmark datasets and three TextZoom low-resolution datasets. Reproduced DAN is slightly different from the open-source model by using data augmentation, bidirectional decoding, etc. With the help of the data augmentation, the recognition accuracy in TextZoom has improved from 33.2\%, 45.2\%, 62.3\% to 36.3\%, 50.0\%, 70.3\%. To some extent, data augmentation improves the generalization in text recognition tasks.

We further implement an ablation study to measure the effectiveness of the fusion module. We discuss the performance from two aspects: training with or without fusion module; different fusion stages of the STR backbone. When we don't use the iterative strategy, IFR$_1$ without fusion module is equivalent to an independent recognition branch. As depicted in Tabel \ref{tab1}, training with fusion module is useful which boosts the accuracy on most of the benchmarks, especially on TextZoom subsets. When training the IFR$_2$ with one iteration, it achieves competitive advantage in accuracy. Besides, IFR$_2$ shows better performance especially on challenging datasets such as SVTP and TextZoom. Besides, we further explore the fusion stage of the STR backbone. The average accuracy on benchmark and TextZoom is in Tabel \ref{tab2} suggests that the deeper the feature depth, the higher the accuracy on test datasets. 

From the fusion module ablation study we can conclude: 1) learning with gradient from the IR branch can make the feature extractor more robust on image quality. 2) by further equipping fusion module with iterative training, the image quality problem can be alleviated, which is recommended to deal with challenging datasets such as TextZoom and SVTP.
\begin{table*}
\caption{Ablation study of fusion stage of the STR backbone.}
\vspace{-0.4cm}
\label{tab2}
\begin{center}
\setlength
\tabcolsep{10pt}
\begin{tabular} {|c|c|c c|}
\hline
Method & Fusion Stage &  Benchmark & TextZoom\\
\hline
IFR$_1$ & 1,2,3 & 87.60 & 53.99 \\
IFR$_1$ & 2,3,4 & 87.69 & 53.95 \\
IFR$_1$ & 1,3,5 & \textbf{87.80} & \textbf{54.31}\\
\hline
\end{tabular}
\end{center}
\end{table*}

\begin{table*}
\caption{Comparison with SOTA methods. 'A' means using data augmentation when training. 'T' means the TextZoom training dataset. Top accuracy for each benchmark is shown in  \textbf{bold}.}
\vspace{-0.4cm}
\label{tab4}
\begin{center}
\begin{tabular} {|c|c|c c c c c c c c|}
\hline
\multirow{2}*{Method} & \multirow{2}*{Training data} &  IIIT & SVT & IC03 & IC13 & IC15 & IC15 & SVTP & CUTE\\
& &  3000   & 647 & 867  & 1015 & 1811 & 2077 & 645  & 288\\
\hline
CRNN(2015) \cite{shi2016end} & MJ &  78.2 & 80.8 & - & 86.7 & -  & -  & - & - \\
FAN(2017) \cite{cheng2017focusing}  & MJ+ST & 87.4 & 85.9 & 94.2 & 93.3 & - & 70.6 & - & - \\
ASTER (2019) \cite{shi2018aster} & MJ+ST & 93.4 & 89.5 & - & 91.8 & 76.1 & - & 78.5 & 79.5 \\
SAR (2019) \cite{li2019show}  & MJ+ST & 91.5 & 84.5 & - & 91.0 & - & 69.2 & 76.4 & 83.3 \\
MORAN (2019) \cite{luo2019moran}  & MJ+ST & 91.2 & 88.3 & 95.0 & 92.4 & - & 68.8 & 76.1 & 77.4 \\
PlugNet(2020) \cite{eccv2020plugnet} & MJ+ST & 94.4 & \textbf{92.3} & 95.7 & \textbf{95.0} & 82.2 & - & 84.3 & \textbf{85.0} \\
DAN(2020) \cite{wang2020decoupled} &MJ+ST & 94.3 & 89.2 & 95.0 & 93.0 & - & 74.5 & 80.0 & 84.4\\
\hline

IFR{$_3$} & MJ+ST+A & 94.6 & 91.7 & 95.6 & 94.2 & - & 78.5 & 83.9 & 82.6\\
IFR{$_3$} & MJ+ST+A+T & \textbf{94.9} & 92.0 & \textbf{96.0} & 94.8 & - & \textbf{80.2} & \textbf{85.4} & 82.3\\
\hline
\end{tabular}
\end{center}
\vspace{-0.7cm}
\end{table*}

\subsection{Comparisons with State-of-the-Arts}
We compare our proposed IFR with state-of-the-art STR methods. Tabel \ref{tab4} shows the recognition results among 7 widely used benchmarks. It is noteworthy that IFR with two iterations outperforms baseline method DAN by a large margin without changing the STR structure. By the progressive collaboration between the STR and IR processes, IFR can help the STR network obtain more robust feature maps and generate higher-quality text images at the same time. Especially in two low-quality text datasets as SVT and SVTP, our method shows a much robust performance. So, the iterative collaboration and RRF may also be useful for other STR networks. When the real low-quality dataset TextZoom adds to training data, the accuracy will be further improved.

Tabel \ref{tab5} shows the results on low-resolution dataset TextZoom. Compared with TSRN, we achieve comparable results with only the synthetic training dataset. Meanwhile, the proposed method achieves better performance when training on both synthetic and TextZoom datasets.

\begin{table}[ht]
\caption{Comparison with SOTA methods on the TextZoom dataset. In each sub--set, the left column only contains 36 alphanumeric characters while the other contains 93 classes, in corresponding to 10 digits, 52 case sensitive letters, 31 punctuation characters. We use the model published on the \href{https://github.com/JasonBoy1/TextZoom/issues/25}{TSRN} to reproduce the results on the 36--class subsets.}
\vspace{-0.4cm}
\label{tab5}
\begin{center}
\begin{tabular}{|c|c|cc|cc|cc|}
\hline
\multirow{2}{*}{Methods} & \multirow{2}{*}{Training Dataset} & \multicolumn{2}{c|}{Hard} & \multicolumn{2}{c|}{Medium} & \multicolumn{2}{c|}{Easy} \\
                         &                                   & 1175        & 1343       & 1209         & 1411        & 1442        & 1619        \\ \hline
\multirow{2}{*}{TSRN}    & Synthetic                         & -           & 33.00    & -            & 45.30     & -           & 67.50     \\
                         & TextZoom                          & 41.45     & 40.10  & 58.56           & 56.30     & 73.79           & 75.10     \\ \hline
\multirow{2}{*}{IFR{$_3$}}   & Synthetic                         & 41.87     & -          & 55.91      & -           & 77.12     & -           \\
                         & Synthetic+TextZoom                          & \textbf{53.19}     & -          & \textbf{69.73}      & -           & \textbf{83.01}     & -           \\ \hline
\end{tabular}
\end{center}
\vspace{-1cm}
\end{table}

\section{Conclusion}
In this paper, we have proposed IFR which explores iterative collaboration for utilizing image restoration knowledge in scene text recognition. The IFR can extract robust feature representation by fusion module RRF and provide refined text images as input for better recognition results by iterative collaboration manner. Quantitative and qualitative results on standard benchmarks and low-resolution dataset TextZoom have demonstrated the superiority of IFR, especially on low-quality images. We also claim that exploiting synthetic augmented data can achieve comparable results.

\bibliographystyle{splncs04}
\bibliography{ref}

\end{document}